\documentclass[hyphens]{article}
\usepackage[final]{neurips_2025} 

\usepackage[utf8]{inputenc} 
\usepackage[T1]{fontenc} 
\usepackage{hyperref} 
\usepackage{subcaption}
\hypersetup{
 colorlinks=true,
 linkcolor=red,
 citecolor=cyan,
 filecolor=magenta, 
 urlcolor=magenta,
 }
 
\usepackage{url} 
\usepackage{booktabs} 
\usepackage{amsfonts} 
\usepackage{nicefrac} 
\usepackage{microtype} 
\usepackage[dvipsnames]{xcolor}
\usepackage{amsmath}
\usepackage{xspace} 
\usepackage{enumitem} 
\usepackage{textcomp}
\usepackage{stfloats}
\usepackage{verbatim}
\usepackage{wrapfig}
\usepackage{graphicx}
\usepackage{subcaption}
\usepackage{amssymb}
\usepackage{cite}
\usepackage{tikz}
\usepackage{algorithm}
\usepackage{algpseudocode}

\usepackage{multicol,multirow} 
\usepackage{threeparttable} 
\usepackage{pgfplots}
\usepackage{adjustbox}
\usepackage{soul}
\usepackage{fancyhdr}
\renewcommand{\headwidth}{\textwidth}

\renewcommand{\headrulewidth}{0.5pt}
\renewcommand{\headrule}{\vspace{2pt}\hbox to\headwidth{\color{black}\leaders\hrule height \headrulewidth\hfill}}
\pgfplotsset{compat=1.18}

\newcommand\modelname{X-Mind}
\title{X-Mind: Efficient Visual Chain-of-Thought via Predictive World Model for End-to-End Driving}
\author{\colorbox{white}{\includegraphics[height=0.8em]{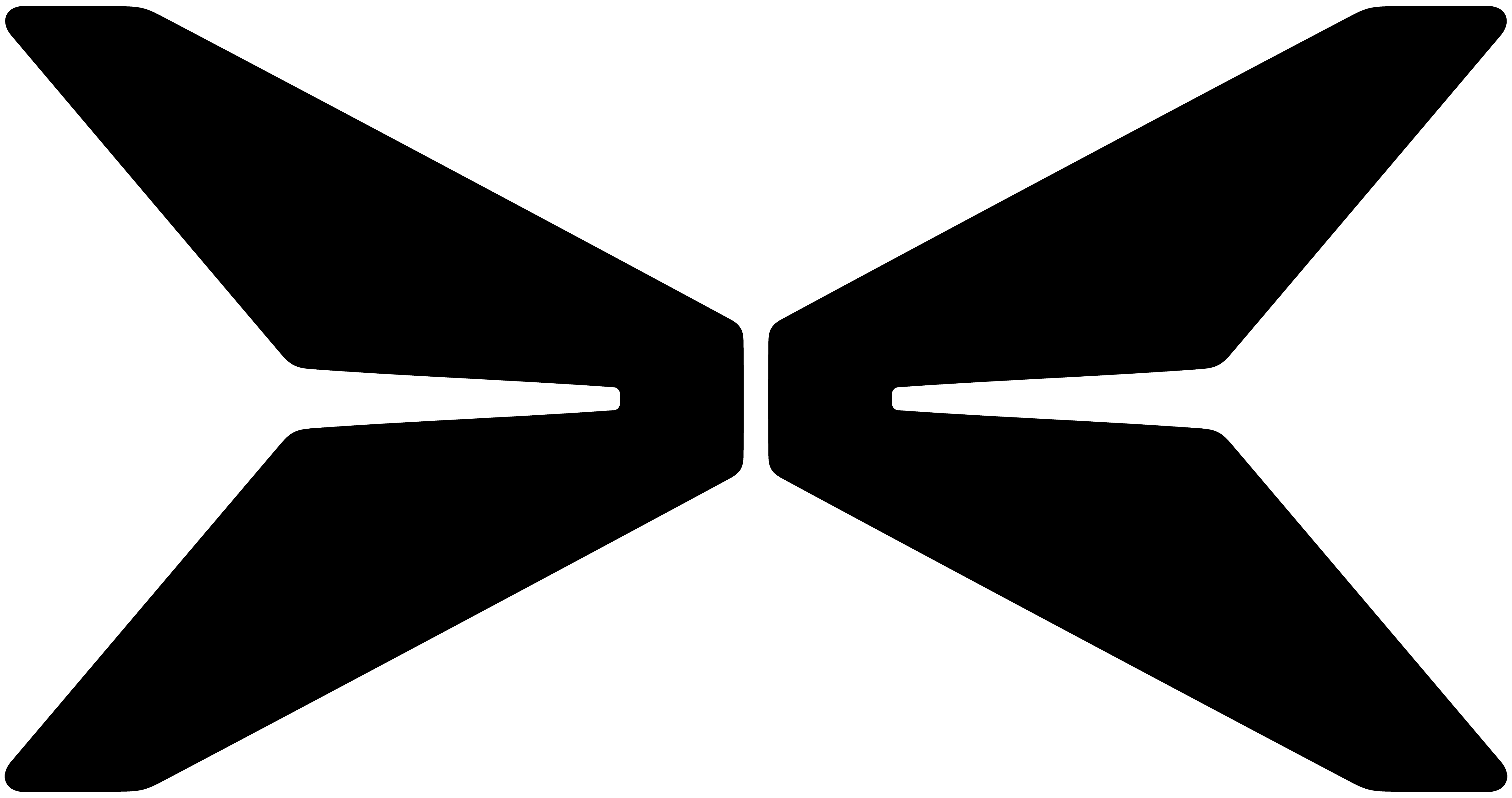}} PWM Team, XPeng Inc. \\
\textbf{\href{https://xp-x-mind.github.io}{\textcolor{NavyBlue}{https://x-mind.github.io}}}} 

\begin{document}
\maketitle

\begin{abstract}
Predicting future states is the cornerstone of reasoning for autonomous physical agents. Although Vision-Language-Action (VLA) models have driven remarkable progress in autonomous driving, they predominantly rely on direct perception to action mapping and fundamentally lack this essential predictive capability. While integrating Predictive World Models (PWMs) has emerged as a consensus to close this gap, existing approaches either cascade PWMs, incurring prohibitive on-vehicle latency, or append them as shallow terminal tasks, failing to fundamentally instill forward-looking reasoning into the deep backbone of large language models.
To endow VLA models with this reasoning capability, we propose \textbf{\modelname}. Rather than treating PWMs as an external auxiliary module, this framework internalizes them as the Visual Chain-of-Thought (Visual CoT). By enforcing a world rollout prior to action, the model is constrained to imagine future evolution first, yielding a driving policy that is robustly grounded in environmental dynamics and aware of the future consequences its actions will unfold.
The challenge here is efficiency, and we tackle it on two fronts. First, we introduce a compact representation of visual thinking: an abstract sketch that fuses a Bird's-Eye-View (BEV) layout with abstract driving priors (e.g., navigation intents and traffic rules). Rather than rolling out dense future frames, the model reasons over this sketch as a mental canvas; aided by a Deep Compression Autoencoder (DC-AE), a 12-frame future rollout is reduced to merely 96 tokens, alleviating the long-context computational bottleneck. Second, to accelerate generation further, we propose a recurrent block diffusion scheme that unrolls the denoising steps across the layers of the large drive model, folding iterative refinement into the backbone's one forward pass.
Trained and validated on large-scale real-world data, X-Mind achieves competitive end-to-end driving performance, which makes it a highly practical, low-latency solution that successfully deploys large-scale cognitive reasoning directly onto resource-constrained vehicle platforms.
\end{abstract}

\section{Introduction}
Reasoning in physical agents fundamentally requires predicting the future because causality unfolds strictly along the temporal axis. The consequences of any immediate decision manifest only in subsequent temporal states. To navigate complex dynamic environments, an autonomous agent must simulate these future evolutions to evaluate its potential actions. Vision-Language-Action (VLA) models~\cite{2025DriveLM, 2024DriveGPT4, Wang2024OmniDriveAH, Fu2025OrionAH, Zhou2025AutoVLAAV, Tian2024DriveVLMTC, Li2025ReCogDriveAR} have advanced end-to-end autonomous driving significantly. However, current architectures primarily rely on reactive mapping from visual perception to control outputs. They process transient environmental features but lack the explicit cognitive capability to anticipate the spatiotemporal evolution of the physical world. Anticipating such evolution requires constructing an internal representation of environmental dynamics to extrapolate how current scenes will unfold~\cite{Zheng2026XWorld,Wang2024DriveDreamer,Gao2023MagicDriveSV,Wen2023PanaceaPA,Yan2024DrivingSphereBA, lu20254d}. 
Concurrent advancements have successfully explored this generative paradigm across diverse application frontiers. Specifically, X-World~\cite{Zheng2026XWorld} introduces a highly controllable and multi-camera generative world model tailored for scalable closed-loop simulation. Concurrently, \mbox{X-Foresight~\cite{Li2026XForesightAJ}} successfully integrates predictive world modeling directly into VLA architectures to jointly learn complex physical dynamics and action control through high-fidelity video forecasting. These complementary efforts establish that integrating Predictive World Models (PWMs) provides a critical pathway for embedding this forward-looking reasoning and instantiating a formal Visual Chain of Thought (Visual CoT).

Deploying PWMs as an internal Visual CoT onboard presents severe computational and architectural challenges. Traditional cascaded paradigms separate future prediction from downstream planning~\cite{Wang2023DrivingIT}. This separation disrupts feature propagation and introduces excessive inference latency. Alternatively, end-to-end joint training approaches \cite{Li2025DriveVLAW0WM,Li2026SGDriveSH} typically append auxiliary reconstruction tasks at the terminal layers of the network. With such appended designs, the supervisory signals generated by the terminal reconstruction tasks are difficult to effectively backpropagate to the deep layers of the heavily parameterized language model backbone. Consequently, the core layers still rely on sparse ground truth trajectories. The model remains susceptible to shortcut learning rather than acquiring genuine physical reasoning capabilities. Solving this cognitive gap requires an internalized world model that is both deeply integrated and highly efficient.

To address this efficiency bottleneck, our primary contribution is the formulation of a visual thinking representation. Cognitive science research, including the foundational theories in image and mind~\cite{1980Image}, indicates that biological intelligence relies on structural and semantic abstractions rather than photorealistic rendering for efficient spatial reasoning. Inspired by this mechanism, we propose an abstract sketch as the intermediate mental canvas for the model. This sketch fuses a Bird's-Eye-View (BEV) layout with abstract driving priors, including navigation intents and traffic rules. By shifting the prediction target from dense future frames to this structured topological representation, we fundamentally reduce the information density required for temporal rollout. Supported by a Deep Compression Autoencoder (DC-AE), this abstract sketch is subsequently compressed into a compact latent manifold. A 12-frame future rollout is efficiently encoded into merely 96 tokens, comprehensively resolving the long-context computational bottleneck.

To accelerate the generation process of this mental sketch, we introduce a Recurrent Block Diffusion (RBD) scheme. Conventional flow matching or diffusion models require repeated forward passes, which is computationally expensive for real-time applications. Our scheme internalizes the generative process by unrolling the progressive denoising steps across the hierarchical layers of the Large Language Model (LLM). This design folds the iterative refinement into a single forward pass of the backbone network. The model directly predicts the future evolution within its latent space. Conditioned on this anticipated physical future, the optimal ego vehicle trajectory is derived through an inverse dynamics planner. This tightly coupled architecture ensures kinematic compliance and provides a robust feature foundation tailored for resource-constrained vehicle platforms.

In summary, the primary contributions of this work are highlighted as follows:

\begin{itemize}
    \item \textbf{Predictive Reasoning via Visual CoT:} We frame physical reasoning as future world prediction, introducing \textbf{{\modelname}} to integrate PWMs as an explicit Visual CoT, providing dense, physics-grounded constraints for VLA models.
    \item \textbf{Efficient Visual Thinking Representation:} Inspired by biological mental imagery, we introduce an abstract sketch representation. This approach fundamentally minimizes the necessary token context for future imagination prediction, effectively resolving the computational bottleneck of instantiating a Visual CoT via PWMs.
    \item \textbf{Recurrent Block Diffusion Scheme:} We design an internalized generative mechanism that unfolds across LLM layers. It achieves extremely fast, single-forward-pass future generation, making high-frequency cognitive reasoning computationally viable for real-time inverse dynamics planning.
\end{itemize}

\section{Method}

\begin{figure*}[ht] 
  \centering
   \includegraphics[width=1.0\linewidth]{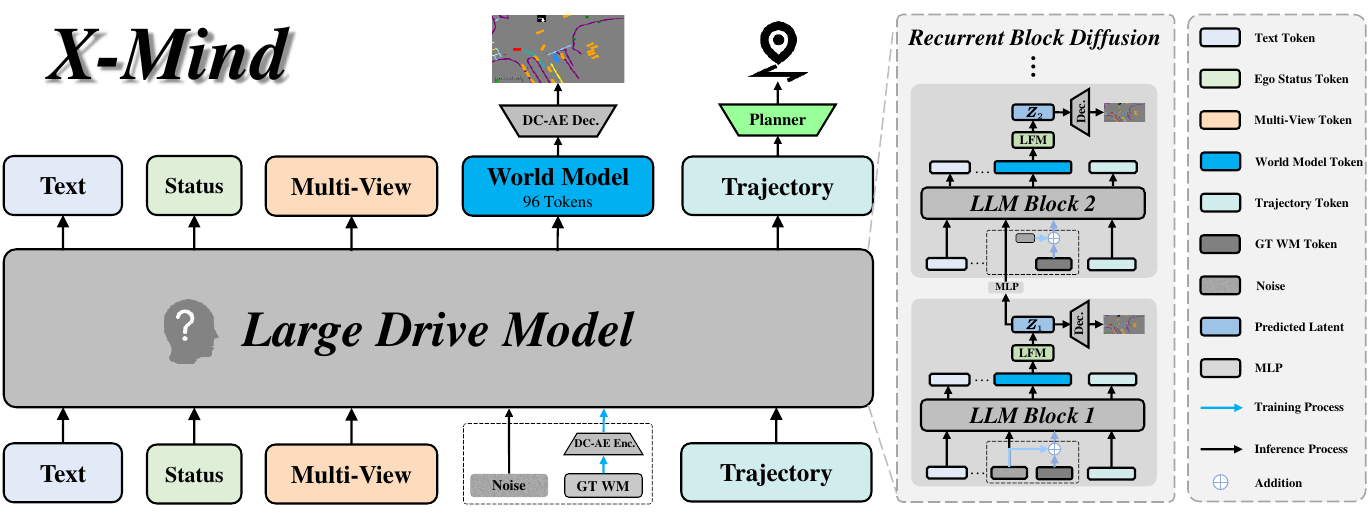} 
   \caption{\textbf{Overall architecture of \modelname.} The PWM is seamlessly embedded within the large drive model. By utilizing Recurrent Block Diffusion, the network executes progressive denoising steps across its hierarchical internal layers in a single forward pass to generate a compact abstract sketch. Conditioned on this anticipated physical future, the planner derives the optimal ego vehicle trajectory. Blue arrows denote the training data flow, while black arrows illustrate the inference process.}
   \label{fig:overview}
\end{figure*}

\subsection{Overview}
Existing VLA models for autonomous driving predominantly follow a direct perception-to-action mapping paradigm. By directly regressing control actions from current visual perception, these models inherently lack the cognitive capacity to explicitly anticipate the future spatiotemporal evolution of the complex physical world. Relying solely on sparse ground truth (GT) trajectories as supervisory signals makes them highly susceptible to shortcut learning. Furthermore, forcefully appending traditional world models as auxiliary terminal tasks often results in severe gradient conflicts within deep networks and introduces long context computational bottlenecks that prohibit real-time onboard deployment.

To systematically address these cognitive and computational limitations, we propose \textbf{\modelname}, as illustrated in Fig.~\ref{fig:overview}. The core philosophy of this framework is to shift from reactive black box mapping to predictive cognitive reasoning. By internalizing a PWM within the large drive model, we instantiate a Visual CoT. This mechanism enforces an explicit spatiotemporal rollout prior to action generation, providing dense and physics-grounded constraints for the deep network features.

To ensure high efficiency during this reasoning process, the architecture relies on two key designs perfectly aligned with the generative data flow. First, it encodes heterogeneous inputs into an efficient visual thinking representation. Rather than predicting dense future images, the model forecasts a compact abstract sketch. It explicitly aggregates fundamental physical scene elements, such as road topologies and dynamic agents, with abstract driving priors. These essential priors incorporate dynamic traffic light states, adaptive navigation intents, and velocity compliance profiles into a unified spatial grid. Supported by DC-AE~\cite{Chen2024DCAE}, this structured representation drastically reduces the token sequence length to merely 96 tokens, resolving the memory constraints.

Second, the framework utilizes an RBD mechanism to accelerate the generation of this future representation. Instead of relying on an external denoising network, this scheme unrolls the progressive denoising steps across the hierarchical internal layers of the large drive model (e.g., LLM Block 1 and LLM Block 2). As indicated by the training and inference pathways in the diagram, the model predicts the anticipated physical future in a single forward pass. Finally, conditioned on this internalized spatial and temporal rollout, a dedicated planner head derives the optimal ego vehicle trajectory via inverse dynamics, guaranteeing kinematically feasible and safe driving behaviors.

\subsection{Visual Thinking Representation}

The core of our framework is an explicit visual thinking representation: rather than reasoning over raw sensory inputs or photorealistic reconstructions, the model imagines the future on an abstract sketch that serves as its mental canvas. We deliberately avoid two common alternatives. Raw perspective camera images suffer heavily from scale variation, occlusion, and depth ambiguity, forcing the model to disentangle geometry before it can reason about dynamics. Continuous 3D reconstructions (e.g., 3D Gaussian Splatting, 3DGS) faithfully recover photometric detail but incur substantial rendering overhead that is largely irrelevant to motion planning. The abstract sketch instead keeps only what matters for driving: it filters out visually irrelevant texture while explicitly preserving the spatial relationships and semantic priors a policy must act upon, achieving an optimal trade-off between information density and computational efficiency.

The sketch is composed of two complementary layers. The first captures the physical environment, rendered from a BEV perspective that projects the scene into a unified, geometrically consistent metric coordinate system. The second goes beyond physical geometry to encode essential driving priors (e.g., navigation intents and traffic rules), which are rasterized onto the very same canvas. By unifying perception and abstract intent in one spatial tensor, the sketch supplies the complete context required for downstream inverse dynamics planning.

\begin{figure}[t]
    \centering
    \includegraphics[width=0.6\linewidth]{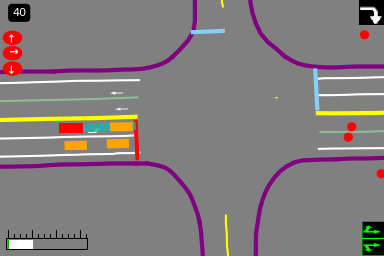} 
    \caption{\textbf{Illustration of the abstract sketch for visual thinking.} Distinct from conventional occupancy maps, our abstract sketch aggregates physical scene elements (dynamic agents and static topologies rendered in a BEV format) with essential driving priors. These incorporated priors explicitly include traffic light states at the top left, adaptive macroscopic navigation paths highlighted as cyan regions, and velocity compliance profiles at the bottom left. Within the structured velocity bar, a green segment represents the actual vehicle speed, a white segment indicates the remaining margin before reaching the speed limit, and a red segment highlights the extent of any velocity exceedance. This unified metric space provides the comprehensive spatial and semantic context fundamentally required for accurate inverse dynamics planning.}
    \label{fig:bev_representation}
\end{figure}

Beyond compactness, the abstract sketch offers strong interpretability. By projecting all objects, map elements, and priors into a single top-down topology, developers can directly diagnose obstacle occupancy, lane connectivity, and multi-agent interactions. This property is especially valuable for LLM-based driving systems: Since transformer architectures are inherently more effective at processing structurally regular information, the sketch substantially reduces the reasoning burden caused by visual noise.

Concretely, distinct from conventional BEV representations that focus narrowly on physical obstacle occupancy or basic object detection, our sketch aggregates both physical environmental states and abstract driving priors into a single comprehensive canvas. As illustrated in Fig.~\ref{fig:bev_representation} and further visualized in Sec.~\ref{sec:more_gt_bev}, it encapsulates the following heterogeneous elements within a unified global context:
\begin{itemize}
    \item \textbf{Physical Scene Elements:} The geometric baseline of the physical world, rendered in BEV, including the ego vehicle (red rectangle), surrounding dynamic agents (yellow/blue rectangles), and explicit road topologies (e.g., lane markings and strict purple boundaries).
    \item \textbf{Dynamic Traffic Light States:} Intersection control states are rasterized directly into the canvas (top-left signal panel), enabling the predictive world model to accurately deduce right-of-way compliance and ego starting and stopping.
    \item \textbf{Adaptive Navigation Intents:} Macro-level routing priors are embedded into the topological space. By rendering the dynamic target path and the ego route (cyan routing region and navigation arrows), the spatiotemporal rollout is constrained to align with macroscopic routing commands.
    \item \textbf{Velocity Compliance Profiles:} Speed-limit semantics and ego-status indicators are integrated into the canvas (bottom-left speed bar), allowing the model to visually enforce legal and safe kinematic execution.
\end{itemize}

Despite its abstraction, the sketch fundamentally remains a dense spatial grid, and directly predicting its pixel-level features within the deep transformer blocks of an LLM would introduce severe sequence-length bottlenecks. To resolve this, we introduce an extreme spatiotemporal compression mechanism. Drawing inspiration from the DC-AE, we train a domain-specific DC-AE tailored exclusively to our structured sketch space. Its encoder compresses the dense, multi-channel sketch tensor into a highly compact, low-dimensional token manifold, reducing a 12-frame future rollout to merely 96 tokens. This compression mitigates the long-context bottleneck of multimodal LLM.

\subsection{Recurrent Block Diffusion}

To enable real-time future prediction without the prohibitive computational overhead of external iterative diffusion models, we introduce the RBD mechanism. This architecture fundamentally relies on a unified Layer Flow Matching (LFM), which internalizes the generative process by unfolding the denoising steps directly across the hierarchical depth of the large drive model. The formulation of this core strategy is detailed across two subsequent subsections. First, we elaborate on the latent rollout process utilized during training to strategically inject progressive noise across specific transformer blocks. Second, we describe the multi-layer velocity prediction scheme that executes efficient Euler integration during inference. Through this unified approach, the model seamlessly constructs the anticipated spatiotemporal states within a single forward pass.

\begin{figure*}[ht] 
  \centering
   \includegraphics[width=0.8\linewidth]{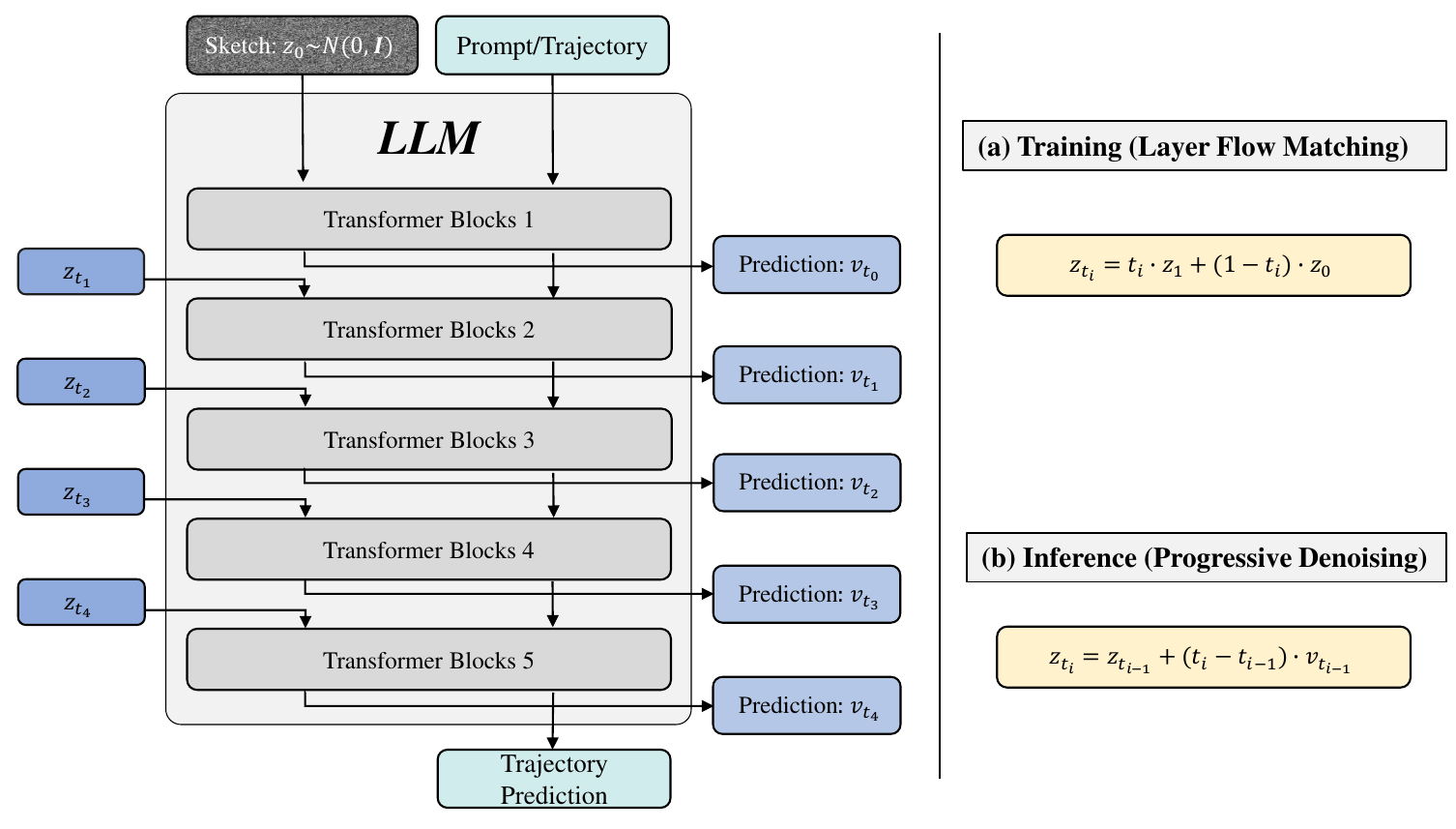} 
   \caption{\textbf{Overview of the Recurrent Block Diffusion. } We first divide all transformer layers of the LLM into five blocks. During training, we replace the sketch-corresponding token features input to each block with the linear combination of noise of different intensities and GT. Then, we extract the sketch-corresponding token features from the outputs of the corresponding blocks and decode them to predict velocities. During inference, we adopt the outputs of the preceding blocks as inputs for the computation of subsequent blocks via Euler integration with a fixed time step.}
   \label{fig:fm}
\end{figure*}

\subsubsection{Layer Flow Matching for Latent Rollout}

In standard continuous-time flow matching~\cite{Lipman2022FM}, an independent neural network (e.g., U-Net~\cite{Ronneberger2015UNetCN} or DiT~\cite{Peebles2022dit}) is iteratively queried to learn a velocity field $\mathbf{v}_\theta\left(\mathbf{z}_t, t\right)$, transporting samples along a linear path from a pure noise distribution to the data distribution:
\begin{equation}
    \mathbf{z}_t = \left(1 - t\right) \cdot \boldsymbol{\epsilon} + t \cdot \mathbf{z}_1, \quad t \in \left[0, 1\right].
\end{equation}
However, executing $N$ iterative forward passes through a massively parameterized denoising network incurs prohibitive computational latency, rendering it unsuitable for real-time autonomous driving.

Our core insight is that: (1) Compared with RGB natural images, abstract sketch eliminates a large amount of high-frequency information that contributes little to autonomous driving while retaining abundant semantic and structural information. Therefore, we argue that elaborately designed complex diffusion models are unnecessary for sketch generation. (2) The varying depths of an LLM naturally encode different levels of abstraction—shallow layers capture low-level patterns, while deep layers encode high-level semantics. This hierarchical structure exhibits a strong isomorphism with the progressive coarse-to-fine denoising process. Leveraging these properties, we propose LFM, which distributes the denoising steps across the internal layers of the LLM. This paradigm achieves a continuous spatiotemporal rollout within a single LLM forward pass, utilizing natural intra-layer token interactions with near-zero additional computational overhead.

During the forward propagation of the LLM, the hidden states of the sketch tokens are explicitly replaced with noisy latent representations at specific layer depths. We define a non-uniform timestep schedule $\left\{t_k\right\}_{k=0}^5 = \left\{0, 0.1, 0.2, 0.4, 0.7, 1.0\right\}$, which maps to specific injection layers in the LLM. At each designated injection layer $l_{\text{inject}}$, the hidden states of the sketch tokens are re-injected as a linear interpolation between noise and the target distribution:
\begin{equation}
    \mathbf{h}^{\left(l_{\text{inject}}\right)}_{\text{bev}} = \text{EncProj}\left(\left(1 - t_k\right) \cdot \boldsymbol{\epsilon} + t_k \cdot \mathbf{z}_1\right) + \text{PE}_{2\text{D}},
\end{equation}
where $\boldsymbol{\epsilon} \sim \mathcal{N}\left(0, \mathbf{I}\right)$ is the initial Gaussian noise, and $\mathbf{z}_1 = \text{Enc}_{\text{DC-AE}}(\mathbf{B}_{\text{gt}})$ represents the target latent representation compressed from the GT future sketch $\mathbf{B}_{\text{gt}}$. $\text{EncProj}(\cdot)$ is a linear projection layer that aligns the dimension of $\mathbf{z}_t$ with the LLM's latent dimension, and $\text{PE}_{2\text{D}}$ denotes a learnable 2D positional encoding.

Through this mechanism, the LLM observes future sketch representations at progressively decreasing noise levels as depth increases. This allows the surrounding perception and trajectory tokens to naturally attend to a clarifying physical future, effectively instantiating the Visual CoT within the latent space.

\subsubsection{Multi-Layer Velocity Prediction and Inference}
To predict the evolutionary velocity field, we extract the hidden states of the BEV tokens from a designated set of layers $\left\{l_k\right\}_{k=1}^5$. At each extraction layer $l_k$, the hidden states are transformed into a velocity prediction:
\begin{equation}
    \mathbf{v}_k = \text{TransEnc}\left(\text{Proj}\left(\mathbf{h}^{\left(l_k\right)}_{\text{bev}}\right)\right),
\end{equation}
where $\text{Proj}$ denotes a linear projection mapping the features back to the latent visual dimension, and $\text{TransEnc}(\cdot)$ is a lightweight transformer encoder whose parameters are strictly shared across all denoising steps to minimize memory overhead.

During inference, the model initiates the rollout from pure noise $\mathbf{z}_0 \sim \mathcal{N}(0, \mathbf{I})$. The latent representation is progressively refined via Euler integration at each specific layer depth:
\begin{equation}
    \mathbf{z}_{k+1} = \mathbf{z}_k + \left(t_{k+1} - t_k\right) \cdot \mathbf{v}_k, \quad k = 0, 1, \ldots, 4
\end{equation}
Following the final update, the fully denoised latent representation $\mathbf{z}_5$ is decoded by a frozen DC-AE decoder to reconstruct the predicted future sketch video sequence.

\subsection{End-to-End Joint Optimization}

The proposed framework is jointly optimized in an end-to-end manner, ensuring both the physical accuracy of the latent world rollout and the kinematic feasibility of the downstream inverse dynamics policy. By unifying these objectives, the continuous latent rollout provides dense, physics-grounded constraints for the LLM backbone, effectively avoiding shortcut learning.

The total loss function $\mathcal{L}_{\text{total}}$ is a weighted sum of the world model rollout loss ($\mathcal{L}_{\text{WM}}$) and the kinematic planning loss ($\mathcal{L}_{\text{plan}}$):
\begin{equation}
    \mathcal{L}_{\text{total}} = \lambda_{\text{WM}} \mathcal{L}_{\text{WM}} + \lambda_{\text{plan}} \mathcal{L}_{\text{plan}}.
\end{equation}

\subsubsection{World Model Rollout Loss}
The internalized world model is optimized with layer-wise latent flow supervision and sparse image-space reconstruction. Instead of training an external denoising network, we supervise the velocity predictions extracted from multiple internal LLM layers. For each selected denoising layer $l_k$, the layer-wise latent flow-matching loss is formulated as:
\begin{equation}
    \mathcal{L}_{\mathrm{flow}}
    =
    \frac{1}{K}
    \sum_{k=1}^{K}
    \left\|
        \mathbf{v}_k-\mathbf{v}^{*}
    \right\|_2^2,
\end{equation}
where $K$ denotes the number of internal denoising layers. This objective encourages all selected LLM depths to learn a consistent transport direction from the noise distribution to the world model latent distribution.

While latent velocity supervision provides an efficient optimization signal, it does not directly constrain the decoded sketch appearance. Using the randomized layer decoding strategy (where only one randomly sampled layer index $r$ is decoded per iteration to minimize computational overhead), we apply a spatial reconstruction loss consisting of pixel-wise mean squared error and perceptual similarity:
\begin{equation}
    \mathcal{L}_{\mathrm{img}}
    =
    \left\|
        \hat{\mathbf{B}}^{(r)}-\mathbf{B}_{\mathrm{gt}}
    \right\|_2^2
    +
    \lambda_{\mathrm{lpips}}
    \mathrm{LPIPS}
    \left(
        \hat{\mathbf{B}}^{(r)},\mathbf{B}_{\mathrm{gt}}
    \right).
\end{equation}

The combined training objective for the world model component is therefore:
\begin{equation}
    \mathcal{L}_{\mathrm{WM}}
    =
    \lambda_{\mathrm{flow}}\mathcal{L}_{\mathrm{flow}}
    +
    \lambda_{\mathrm{img}}\mathcal{L}_{\mathrm{img}}.
\end{equation}
This strategy allows the model to learn a structured BEV latent rollout with minimal additional overhead beyond the original LLM forward pass.

\subsubsection{Kinematic Planning Loss}
Concurrently, the inverse dynamics planning head is supervised by $\mathcal{L}_{\text{plan}}$. Rather than relying solely on standard unconstrained coordinate regression, this loss directly applies L1 supervision to the parameterized kinematic control actions (longitudinal acceleration $a_{\text{lon}}$ and yaw rate $\omega_{\text{yaw}}$). This ensures that the generated trajectories remain physically executable and adhere strictly to the non-holonomic constraints of the ego-vehicle.

\section{Experiments}

\subsection{Datasets}
To validate the effectiveness of the proposed approach, we utilize a large-scale internal autonomous driving dataset collected from diverse real-world scenarios. This comprehensive dataset contains approximately \textbf{280,000 hours} of continuous driving records, which are systematically segmented into \textbf{34M} video clips. Following the hardware configuration established in X-World~\cite{Zheng2026XWorld}, the sensory input comprises multi-view images captured by a suite of seven onboard cameras, providing full 360-degree environmental coverage. These cameras specifically include front fisheye, front narrow, left front, right front, left rear, right rear, and rear perspectives.

We adopt the same data processing protocol as X-Foresight~\cite{Li2026XForesightAJ}. The resulting visual inputs are tokenized into a massive corpus of \textbf{13.8T} tokens to facilitate learning of the large drive model, with the dataset distribution spanning approximately 86.8\% urban and 13.2\% highway driving conditions. This extensive and diverse foundation provides robust supervision for the model to internalize both routine navigation and complex long-tail interactions. All subsequent experiments presented in this work are conducted on a subset comprising one eighth of this total dataset.

\subsection{Impact of Different Scene Representations}
\label{sec:scene_representation_ablation}

To justify selecting the abstract sketch as the intermediate latent target, we compare different scene representations using identical backbones and training configurations. The evaluated candidates include raw image features, 3DGS, and our proposed abstract sketch, which is instantiated in a structured BEV format enriched with explicit driving priors. This experiment systematically evaluates whether an abstract, intent-driven scene representation can improve spatial reasoning oriented towards planning while maintaining practical training and inference efficiency.

As summarized in Tab.~\ref{tab:representation_ablation}, directly introducing future image features improves trajectory prediction compared to the standard base model. However, this naive approach requires a massive number of additional tokens (3584 extra tokens), leading to severe memory constraints and computational burdens for the large drive model. Alternatively, utilizing 3DGS provides continuous spatial geometric information. However, its reconstruction-focused nature introduces substantial token overhead (3072 extra tokens) and yields only marginal planning improvements over the raw image baseline.

In stark contrast, our abstract sketch representation achieves the best Average Displacement Error performance across both lateral and longitudinal dimensions. Crucially, it accomplishes this superior planning accuracy while requiring a mere 96 additional tokens. This represents an extreme reduction in context length compared to the dense image and 3DGS baselines. This phenomenon explicitly demonstrates that the abstract sketch effectively filters out visual details irrelevant to planning. It preserves only the structured spatial priors and essential driving semantics necessary for downstream trajectory prediction.

These quantitative results indicate that the abstract sketch provides the optimal balance between representation compactness, geometric alignment, and computational efficiency.

\begin{table}[htbp]
    \centering
    \caption{Performance and efficiency comparison of different scene representations.}
    \label{tab:representation_ablation}
    \adjustbox{max width=1.0\textwidth}{
    \begin{tabular}{lcccc}
        \toprule
        Method & Extra Tokens  & ADE Lat. $\downarrow$ / ADE Lon. @6s $\downarrow$  & Inference $\downarrow$\\
        \midrule
        Base         & 0     & 0.2399 / 1.2979 & \textbf{1.0}\\
        Base + Image & 3584  & 0.2003 / 1.2456 & 22.0\\
        Base + 3DGS  & 3072  & 0.1964 / 1.2247 & 19.0\\
        Base + Sketch  (Ours)   & 96    & \textbf{0.1765} / \textbf{1.1849} & 1.1\\
        \bottomrule
    \end{tabular}
    }
\end{table}

\subsection{Comparison of Diffusion Architectures}
\label{sec:main_results}

To validate the effectiveness of our proposed framework, we establish a comprehensive comparison against two representative baselines. The first is a standard VLA baseline. This standard architecture directly predicts control trajectories from visual perception, completely omitting explicit future reasoning. The second is a single-step denoising process. Rather than distributing the generative mechanism across multiple intermediate layers, this baseline treats the entire language backbone as a unified generative model to execute a direct one-step velocity prediction. We systematically compare these approaches against our proposed Recurrent Block Diffusion framework, which computes progressive velocity predictions distributed across multiple network blocks, to evaluate both generative visual quality and downstream planning performance.

Tab.~\ref{tab:model_architecture} reports the quantitative comparison across inference latency, future prediction quality (measured by FID), and trajectory planning accuracy (measured by ADE). Integrating a single-step denoising process improves planning performance over the standard VLA baseline, confirming the fundamental necessity of explicit future supervision. However, executing the diffusion process entirely at the final stage yields a remarkably high FID of 67.30, indicating severe modality collapse. In stark contrast, our proposed RBD framework drastically reduces the FID to 9.59. Most notably, the RBD mechanism achieves this immense generative superiority with negligible computational overhead compared to the single-step baseline, introducing a minimal $0.1\times$ inference delay. Supported by this highly efficient and physically accurate future rollout, the model further minimizes both lateral and longitudinal ADE, establishing the state-of-the-art planning performance.

\begin{table}[htbp]
    \centering
    \caption{Performance and efficiency comparison across different architecture paradigms.}
    \label{tab:model_architecture}
    \adjustbox{max width=1.0\textwidth}{
    \begin{tabular}{lccc}
        \toprule
        Method  & Inference$\downarrow$ & FID $\downarrow$ & ADE Lat. $\downarrow$ / ADE Lon. $\downarrow$ \\
        \midrule
        Base                        & \textbf{1.0} & -- & 0.2399 / 1.2979 \\
        Base + Sketch (Single Step)     & 1.1 & 67.30 & 0.1783 / 1.1938 \\
        Base + Sketch (RBD)   & 1.1 & \textbf{9.59} & \textbf{0.1765} / \textbf{1.1849} \\
        \bottomrule
    \end{tabular}
    }
\end{table}

\begin{figure*}[t]
  \centering
   \includegraphics[width=1.0\linewidth]{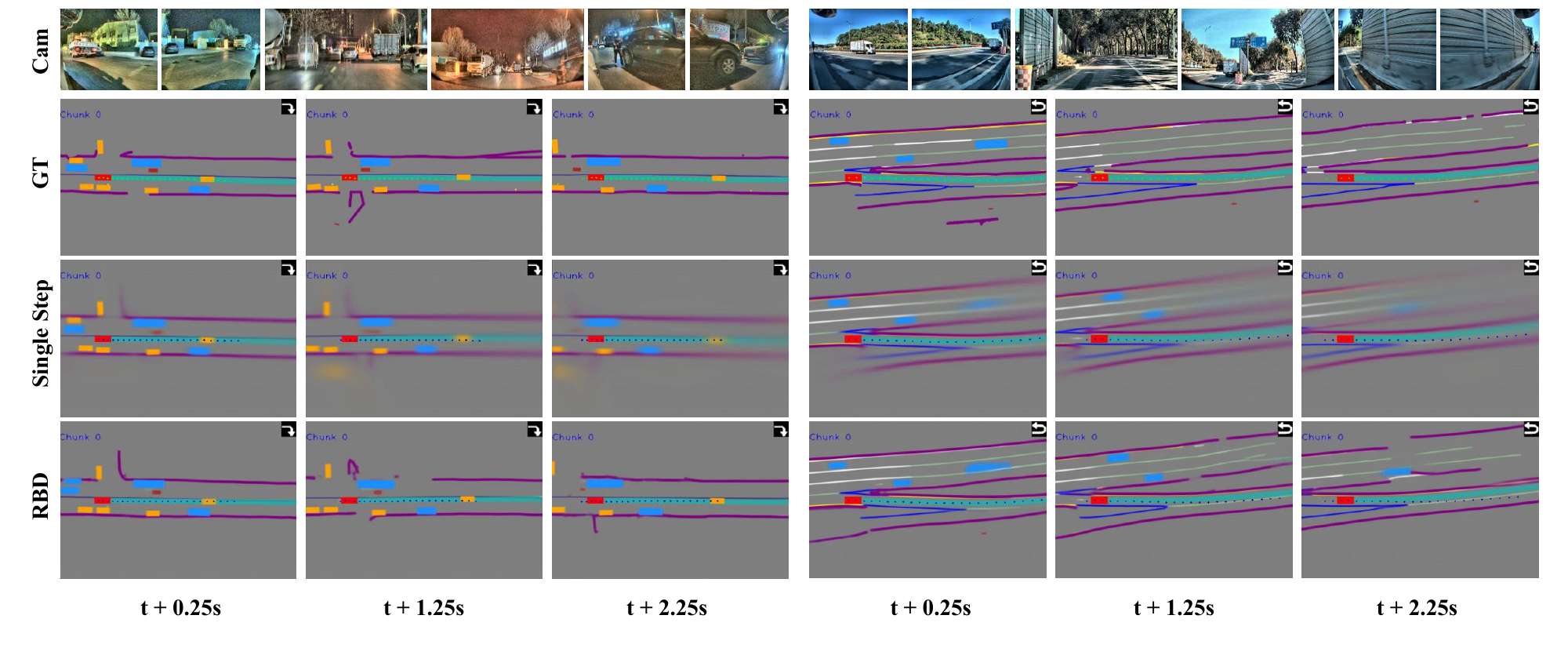}
    \caption{\textbf{Qualitative comparison of future world prediction.} We present spatial rollout results across daytime and nighttime scenarios. The proposed RBD framework (bottom row) produces strictly accurate and temporally coherent predictions compared to the single-based baseline (middle row). Crucially, the RBD framework demonstrates the cognitive capability to forecast the motion of dynamic objects even when they are missing from the ground truth supervision.}
   \label{fig:result}
\end{figure*}

Fig.~\ref{fig:result} provides a qualitative evaluation of the future spatiotemporal predictions across diverse daytime and nighttime driving conditions. Compared to the single-based baseline, which frequently produces blurred or fading semantic layouts over time, the RBD framework generates persistently sharp and temporally coherent abstract sketches. Furthermore, even when specific dynamic agents are missing or heavily occluded in the ground truth annotations, the RBD model successfully infers their continuous motion trajectories based on contextual traffic interactions. This phenomenon demonstrates that the internalized diffusion process acquires a profound understanding of physical dynamics and causal reasoning, rather than merely memorizing the training labels.

\subsection{Reconstruction vs. Future Generation}
\label{sec:reconstruction_vs_generation}

Further analysis investigates whether the fundamental benefit of abstract sketch supervision originates from reconstructing the current observation or from generating future states. To this end, we compare three sketch prediction targets: reconstructing the current sketch frame, predicting one future sketch frame, and predicting a 12-frame future sketch sequence. The first setting corresponds to a reconstruction task, while the latter two require the model to anticipate future scene evolution.

\begin{table}[htbp]
    \centering
    \caption{Ablation study on temporal prediction targets under the proposed framework.}
    \label{tab:bev_temporal_target_ablation}
    \adjustbox{max width=1.0\textwidth}{
    \begin{tabular}{llcc}
        \toprule
        Method & Sketch Target & FID $\downarrow$ & ADE Lat. $\downarrow$ / ADE Lon. $\downarrow$ \\
        \midrule
        Base + Sketch (RBD) & Current frame & \textbf{8.97} & 0.1866 / 1.2132 \\
        Base + Sketch (RBD) & Future 1 frame & 9.05 & 0.1840 / 1.2124 \\
        Base + Sketch (RBD) & Future 12 frames & 9.59 & \textbf{0.1765} / \textbf{1.1849} \\
        \bottomrule
    \end{tabular}
    }
\end{table}

As shown in Tab.~\ref{tab:bev_temporal_target_ablation}, current-frame sketch reconstruction achieves the best FID, indicating that reconstructing the observed scene is visually easier. However, it leads to the weakest trajectory prediction performance among the three settings. This suggests that high-fidelity reconstruction of the current scene alone does not provide sufficient supervision for future motion reasoning.

Predicting a single future sketch frame improves ADE over current-frame reconstruction, despite having a slightly worse FID. This indicates that even short-horizon future generation introduces planning-relevant temporal cues, such as ego-motion-induced spatial shifts, surrounding-agent dynamics, and navigation-conditioned scene evolution. These cues are more directly related to downstream trajectory prediction than static scene reconstruction.

The best trajectory performance is achieved by predicting 12 future frames. Although this setting has a higher FID due to the increased difficulty of long-horizon generation, it provides stronger supervision for modeling temporal scene evolution. The model is encouraged to learn dynamic traffic interactions, future occupancy changes, and long-range navigation consistency, which leads to more planning-aligned representations.

These results reveal that FID and trajectory metrics reflect different aspects of the world model. FID mainly measures image-level distribution quality, whereas ADE reflects downstream planning utility. The current-frame reconstruction setting achieves the lowest FID but inferior ADE, while the 12-frame future generation setting achieves the best ADE with a moderately higher FID. Therefore, we conclude that the core advantage of the world model fundamentally derives from its predictive generative rollout, rather than from visually faithful reconstructions of present observations.

\subsection{Analysis of Structured Sketch Ground Truth}
\label{sec:more_gt_bev}

\begin{figure}[ht]
\centering
\begin{subfigure}{0.6\textwidth}
    \centering
    \includegraphics[width=0.48\linewidth]{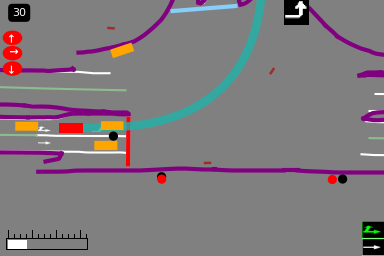}\hfill
    \includegraphics[width=0.48\linewidth]{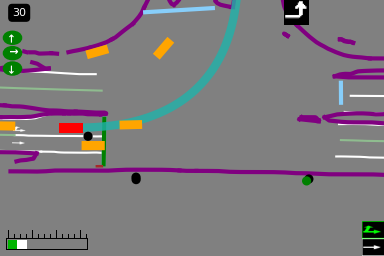}
    \caption{Traffic Light Transition GT ($t=0$s vs. $t=1$s)}
    \label{fig:qual_traffic}
\end{subfigure}

\begin{subfigure}{0.6\textwidth}
    \centering
    \includegraphics[width=0.48\linewidth]{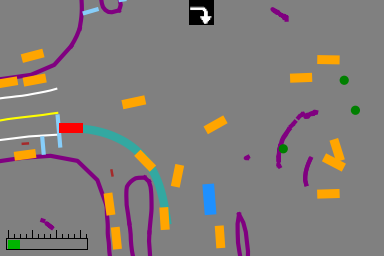}\hfill
    \includegraphics[width=0.48\linewidth]{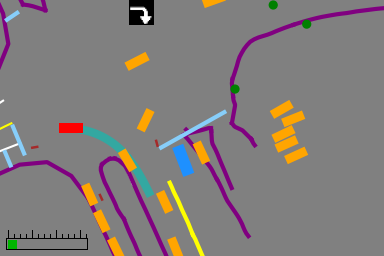}
    \caption{Adaptive Navigation Target Path GT ($t=0$s vs. $t=1$s)}
    \label{fig:qual_nav}
\end{subfigure}

\begin{subfigure}{0.6\textwidth}
    \centering
    \includegraphics[width=0.48\linewidth]{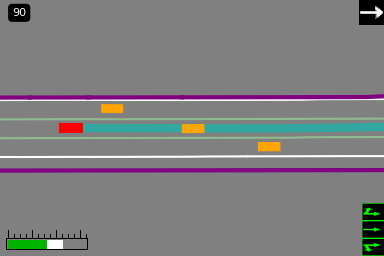}\hfill
    \includegraphics[width=0.48\linewidth]{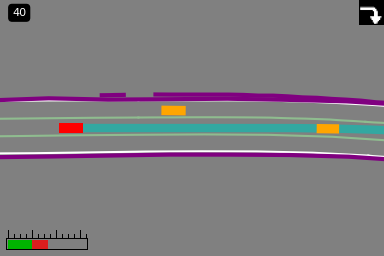}
    \caption{Velocity Compliance Profile GT (Under-speed vs. Over-speed)}
    \label{fig:qual_speed}
\end{subfigure}

\caption{\textbf{Visualization of the structured sketch GT data.} These labels represent the high-fidelity supervision signals used to train the world model, incorporating (a) traffic light signal phases, (b) macroscopic routing intents, and (c) explicit kinematic compliance bars. Such dense and structured labels are crucial for the model to internalize complex physical and semantic driving rules.}
\label{fig:dynamic_rollout}

\end{figure}

To further elucidate the semantic richness and structural complexity of the supervision signals used in our framework, we provide a detailed analysis of the structured sketch GT data. Unlike conventional sparse labels, our GT representation integrates high-fidelity physical layouts with abstract driving priors. As illustrated in Fig.~\ref{fig:dynamic_rollout}, these comprehensive labels provide the necessary spatial and temporal context to supervise the world model's learning of complex driving logic.

\paragraph{Intersection Control and Signal Encoding (Fig.~\ref{fig:dynamic_rollout}a).}
The GT sequences explicitly incorporate traffic light states into the spatial grid to supervise the model's temporal reasoning. In this intersection scenario, the GT data captures the precise transition from a stationary phase ($t=0$s, Stopped) to a dynamic proceeding phase ($t=1$s, Proceeding) following a signal change. By providing such dense supervision, the framework ensures that the world model learns the causal relationship between signal phases and vehicle start-stop mechanics.

\paragraph{Navigation-Guided Routing Priors (Fig.~\ref{fig:dynamic_rollout}b).}
To ground the model's future rollouts in high-level routing intents, the GT data embeds macro-navigation targets directly into the topological space. The visualization of the progressive GT sequence reveals how the designated cyan routing corridor and navigation arrows update adaptively as the ego-vehicle advances. This structured guidance serves as a deterministic spatial constraint, ensuring the model's latent imagination remains strictly aligned with the intended target path.


\paragraph{Kinematic Velocity Compliance (Fig.~\ref{fig:dynamic_rollout}c)} The GT data also provides explicit supervision for speed limit semantics and kinematic states. By comparing the abstract sketch labels across different velocity profiles, the supervision distinctly differentiates between compliant and excessive speed conditions. This kinematic status is visually quantified by a speed bar located at the bottom left of the sketch. Within this structured bar, a green segment represents the actual vehicle speed, a white segment indicates the remaining margin before reaching the speed limit, and a red segment explicitly highlights the extent of any velocity exceedance. By coupling the physical environment with these legal kinematic thresholds, the labels compel the predictive world model to accurately distinguish between safe and noncompliant future states during its continuous spatial rollout.

\subsection{Qualitative Visualiztion}
As shown in Figure 6, we provide some visual examples stemming from internal autonomous driving dataset for X-Mind. By the Figure, X-Mind with RBD mechanism empowered by visual thinking demonstrates excellent and progressive planning results.

\begin{figure}[htbp]
    \centering
    \begin{subfigure}[b]{0.48\textwidth}
        \includegraphics[width=\textwidth]{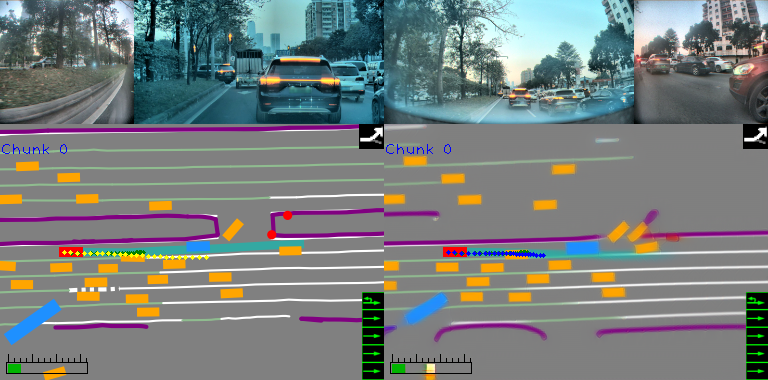}
        \caption{Safety: Frontal Vehicle Sudden Braking}
    \end{subfigure}
    \hfill
    \begin{subfigure}[b]{0.48\textwidth}
        \includegraphics[width=\textwidth]{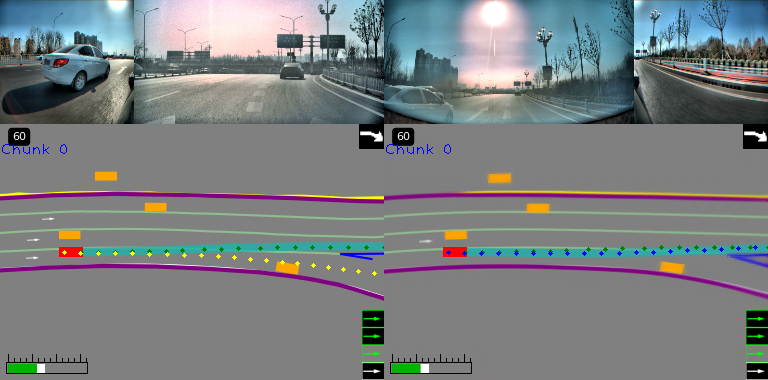}
        \caption{Navigation: Lane Keeping at Off-Ramp}
    \end{subfigure}
    
    \vspace{0.3cm}
    \begin{subfigure}[b]{0.48\textwidth}
        \includegraphics[width=\textwidth]{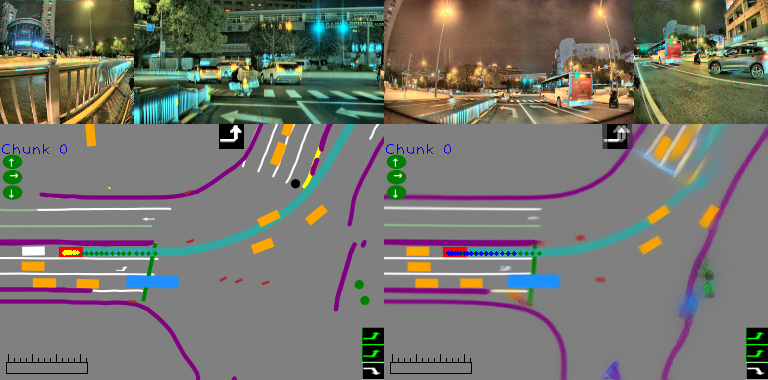}
        \caption{Traffic Lights: Compliance \& Start}
    \end{subfigure}
    \hfill
    \begin{subfigure}[b]{0.48\textwidth}
        \includegraphics[width=\textwidth]{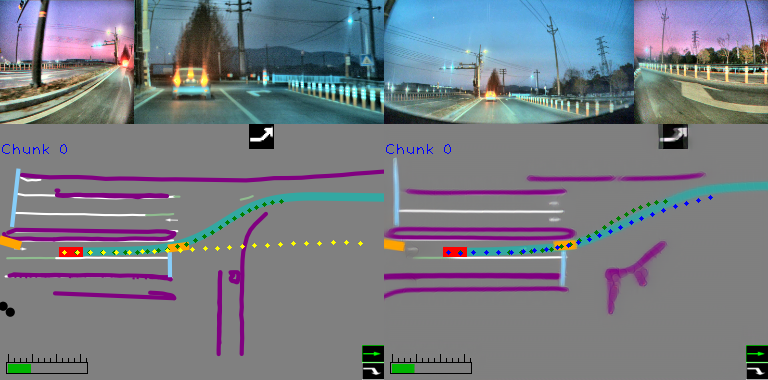}
        \caption{Topo \& Obstacle Avoidance}
    \end{subfigure}

    \vspace{0.4cm}
    \small
    \raisebox{0.5pt}{\textbullet} {\color{green!70!black} Ground Truth} \quad
    \raisebox{0.5pt}{\textbullet} {\color{yellow!90!black} w/o World Model} \quad
    \raisebox{0.5pt}{\textbullet} {\color{blue!80!black} w/ World Model} \\
    \vspace{0.2cm}
    \textit{Layout: Top row shows 4-view cameras; bottom-left: GT BEV; bottom-right: WM-predicted BEV.}
    
    \caption{\textbf{Qualitative comparison of driving trajectories}. The World Model enhances safety, navigation adherence, and semantic understanding in complex urban environments.}
    \label{fig:comparison}
\end{figure}

\section{Future Work}
Despite the advancements introduced by this architecture, we identify two primary directions for ongoing and future research to further enhance the cognitive capabilities and scalability of the system.

First, we are actively exploring the joint sampling of control actions and the intermediate abstract sketch representations. The current inference pipeline sequentially derives the optimal ego vehicle trajectory conditioned on the anticipated physical future. Transitioning to a joint generation process would allow the model to evaluate kinematic feasibility and complex environmental interactions simultaneously during the internal denoising steps. This unified sampling strategy aims to further improve the temporal consistency and safety margins of the planned trajectory under highly interactive and densely populated driving scenarios.

Second, while the current architecture demonstrates exceptional performance trained on a massive internal dataset, its inherent reliance on GT annotations for the structured spatial representations presents a potential bottleneck for continuous scaling. To address this limitation and facilitate exponential model growth, future iterations will focus on integrating self-supervised representation learning paradigms. By training the Visual CoT to forecast future states directly from unannotated sensory inputs or through unsupervised latent dynamics, the framework can leverage virtually infinite amounts of raw driving logs. This shift towards self-supervised learning will significantly ease the effort to scale up the architecture, ultimately unlocking deeper physical intuition and more robust cognitive reasoning capabilities for end-to-end autonomous systems.

\section{Conclusion}
\label{sec:conclusion}

In this report, we present \textbf{\modelname}, a framework that endows VLA models with forward-looking reasoning capabilities. Rather than cascading separate models or appending shallow terminal tasks, the architecture embeds predictive world modeling directly into the deep backbone as a Visual CoT. By enforcing an explicit world rollout prior to action generation, the driving policy becomes robustly grounded in environmental dynamics and fully aware of the future consequences its actions will unfold. 
To systematically overcome the efficiency bottlenecks associated with this cognitive reasoning, we tackle the challenge on two fronts. First, we introduce an extremely compact abstract sketch that fuses BEV layouts with essential driving priors. Supported by a DC-AE, this mechanism compresses the future context into a minimal token footprint, effectively alleviating the long context computational burden. Second, we propose an RBD scheme that accelerates generation by unfolding the iterative denoising steps across the layers of the large drive model, folding the refinement process into a single forward pass. 
Validated on extensive real-world driving data, {\modelname} achieves highly competitive end-to-end performance. Finally, this work establishes a practical and low-latency solution that successfully deploys large-scale cognitive reasoning directly onto resource-constrained vehicle platforms.

\section*{Contributors}
\newcommand{\equalIT}{\textsuperscript{*}}
\newcommand{\internIT}{\textsuperscript{\ensuremath{\ddagger}}} 

We extend our sincere gratitude to the entire team for their dedication and hard work. This project is a testament to our collective effort in pushing the boundaries of world model research and engineering.

\vspace{1em}

\begin{description}
    \item[Advisors:] Yu Zhang, Hang Zhang, Xianming Liu
    \item[Project Lead:]  Qingyu Luo, Zhuangzhuang Ding
    \item[Contributors:] Bohao Zhao\equalIT, Chengrui Wei\equalIT, Guangfeng Jiang\equalIT\internIT, Ruixin Liu\equalIT,  Xuejie Lv\equalIT, Liu Liang, Sutao Deng, Xiuyang Fan, Pengkun Zheng, Jinyun Zhou, Rui Guo, Hanpeng Liu, Yutong Zheng,  Yi Guo, Xinlong Zheng
\end{description}
\textbf{Technical Program Manager}: Tenglong(Victor) Gu

{\footnotesize
\noindent \equalIT Core contribution. The first five authors are listed in alphabetical order. \\
\internIT Research Intern at XPENG.}

{ 
\small
\bibliography{neurips_2025}
\bibliographystyle{plain} 
}

\end{document}